\documentclass[letterpaper]{article}

\usepackage{natbib}

\usepackage{makeidx}  
\usepackage{alltt}
\usepackage{marvosym}

\newif\ifFigs          \Figsfalse
\newif\ifWorkComments  \WorkCommentsfalse

\Figstrue 
\WorkCommentstrue
\usepackage{epsfig,subfigure}
\usepackage{amsmath,amssymb}
\usepackage{float}
\usepackage{rotating}
\usepackage{textcomp}

\usepackage[colorlinks=false]{hyperref}  

\usepackage{graphpap}
\usepackage{color}
\definecolor{darkgreen}{rgb}{0.1, 0.7, 0.0}
\definecolor{darkred}{rgb}{0.5, 0.1, 0.1}

\newcommand{\workComment}[1]%
{\ifWorkComments\begin{quote}\footnotesize\color{darkgreen} #1\end{quote}\fi}

\usepackage{dsfont}

\newcommand{\A}{{\mathcal A}}

\title{Artificial Hormone Reaction Networks:\\Towards Higher Evolvability in Evolutionary Multi-Modular Robotics}

\author{Heiko Hamann, J{\"u}rgen Stradner,\\Thomas Schmickl, Karl Crailsheim\\
Artificial Life Lab of the Department of Zoology,\\Karl-Franzens University Graz, Universit{\"a}tsplatz 2,\\A-8010 Graz, Austria,\\
heiko.hamann@uni-graz.at}

\begin{document}
\maketitle


\begin{abstract}
  The semi-automatic or automatic synthesis of robot controller
  software is both desirable and challenging. Synthesis of rather
  simple behaviors such as collision avoidance by applying artificial
  evolution has been shown multiple times. However, the difficulty of
  this synthesis increases heavily with increasing complexity of the
  task that should be performed by the robot. We try to tackle this
  problem of complexity with Artificial Homeostatic Hormone Systems
  (AHHS), which provide both intrinsic, homeostatic processes and
  (transient) intrinsic, variant behavior. By using AHHS the need for
  pre-defined controller topologies or information about the field of
  application is minimized. We investigate how the principle design of
  the controller and the hormone network size affects the overall
  performance of the artificial evolution (i.e., evolvability). This
  is done by comparing two variants of AHHS that show different
  effects when mutated. We evolve a controller for a robot built from
  five autonomous, cooperating modules. The desired behavior is a form
  of gait resulting in fast locomotion by using the modules' main
  hinges.
\end{abstract}

\section{Introduction}

The (semi-)automatic synthesis of robot controllers with artificial
evolution belongs to the software section of evolutionary
robotics~\citep{cliff93}. The main challenge in this field is the
curse of complexity because an increase in the difficulty of the
desired behavior results in a significantly super-linear increase in
the complexity of its evolution. This is partially documented by the
absence of complex tasks in the
literature~\citep{nelson_2009_fitness}. Additionally, in evolutionary
robotics the cost of the fitness evaluation is rather high even in
case of simulations, if the application of a physics engine
(simulation of friction, inertia etc.) cannot be avoided. Another
challenge is the appropriate choice of a genetic
encoding~\citep{mataric_1996_challenges} and the basic principle of
the controller design as they define the designable fraction of the
search space and the fitness landscape (non-designable fractions are
induced, for example, by the environment or the task itself). While
the search space should be kept small, the fitness landscape should be
smooth with a minimum number of local optima. Experience shows that
these two criteria are contradicting. We summarize this complex of
challenges by the aim to `strive for high evolvability'.

Concerning the problem of finding appropriate controller designs a
pleasant trend can be observed in recent literature. The most
prominent candidate is presumably the HyperNEAT
design~\citep{stanley09,clune09}. It is based on artificial neural
networks (ANN) but combines the `search for appropriate network
weights with complexification of the network
structure'~\citep{stanley04} through the generation of connectivity
patterns. It has proven to have good evolvability combined with an
adequate range of applications. Other promising, recent approaches
tend to be more inspired by biology, in particular by unicellular
organisms and endocrine systems. Examples showing good evolvability
are the reaction-diffusion controller by \citet{dale10} and
homeostasis and hormone systems based on GasNets~\citep{vargas09} and
ANNs~\citep{neal03}. They indicate homeostasis as a prominent feature
in successful adaptation to dynamic environments.

In this paper, we analyze a controller design called Artificial
Homeostatic Hormone Systems (AHHS) that is based on hormones only and
was introduced
before~\citep{hamann10b,schmickl10a,schmickl09a,stradner_2009_evolving,stradner_2009_analysis}. AHHS
is a reaction-diffusion approach. Sensory stimuli are converted into
hormone secretions that, in turn, control the actuators. In addition,
hormones interact linearly and non-linearly comparable to the hidden
layer of ANN. The topology of this hormone-reaction network is not
predefined. Such systems show homeostatic processes because they
typically converge to trivial equilibria for constant sensor
input. The sensory stimuli are basically integrated in form of hormone
concentrations (a form of memory) and decomposed over time
(oblivion). However, during a limited period of time (transient) after
a stimulus they show also variant behavior, especially, if non-linear
hormone-to-hormone interactions are applied. This way, explorative
behavior of the robot is implemented that allows for the testing of
many sensory-motor configurations. The concept of AHHS is related to
gene regulatory networks. However, here each edge has its own
activation threshold and redundant edges with different activations
between two hormones are allowed.

The desired main application of AHHS is multi-modular
robotics~\citep{symbrion,replicator}. In this field, autonomous
robotic modules are studied, that are able to physically connect to
each other, and can also establish a communication and energy
connection. Hence, they form a super-robot called `organism', that is
able to re-configure its body shape, see for example, \citet{shen06}
or \citet{murata08}. Therefore, the underlying idea of diffusion in
our reaction-diffusion system is that hormones diffuse from robot
module to robot module and establish a low-level
communication. Following our maxim of trying to reach a maximum of
plasticity we use identical controllers in each module independent of
their position within the robot organism, so there is neither a
controller nor a module specialization. This concept implements the
focus of evolutionary robotics on modularity (among others) in terms
of hardware and software~\citep{nolfi04}. Although we evolve
cooperative behaviors by evolving a kind of self-organized role
selection, there is no co-evolution.


In general, our approach is more organic in contrast to the typical
symbolic approach (direct encoding of pitch, roll, yaw angles, use of
pattern generators using Gaussian functions etc.). The biological
inspiration is not practiced as an end in itself but rather introduces
more robustness in computations and it allows the diffusion of such
values from module to module (implementing implicit communication).

One focus of our current research track is to design fitness
landscapes by using appropriate controller designs. We investigate
possibilities of smoothing the fitness landscape by a sophisticated
interaction between the controller design and the mutation
operator. We test whether it is useful to maximize the causality of
the mutation operator (i.e., small causes have small effects) by
reducing the maximal impact to the organism's behavior. However,
whether high causality is really desirable, is questionable (e.g.,
cf.~\citet{chouard10}).

The investigated scenario is a modular-robotics variant of gait
learning in simulation. Initially, we connect five modules in a simple
chain formation as the body formation itself is not yet in our
focus. The task is to move as far as possible by utilizing the hinge
in each module only (no wheels).

\section{Artificial Homeostatic Hormone Systems}

In AHHS, sensors trigger hormone secretions, which increase hormone
concentrations in the robot. These hormones diffuse, integrate, decay,
interact and finally, affect actuators. We have analyzed AHHS
controllers in single robots
before~\citep{schmickl10a,schmickl09a,stradner_2009_evolving,stradner_2009_analysis}. In
these cases, the robot's body was virtually divided into compartments
that hold hormones and between which hormones diffuse. These
compartments create a spatial context (embodiment) by associating
sensors and actuators with explicit compartments (e.g., left proximity
sensor and left wheel actuator are associated with the left
compartment and hence depend only on hormone concentrations of this
compartment). In the case of modular robotics, the subdivision of the
robot organism is naturally defined by the modules
themselves. A~virtual compartmentalization is not necessary and
hormones diffuse from module to module (see
Fig.~\ref{fig:ahhsPicto}). A first small case study with organisms
built from three modules was reported in~\citep{hamann10b}.

\setlength{\unitlength}{0.008\textwidth}
\begin{figure}
  \centering
  \begin{picture}(120,40)
    \put(0,0){\includegraphics[angle=0,width=120\unitlength]
      {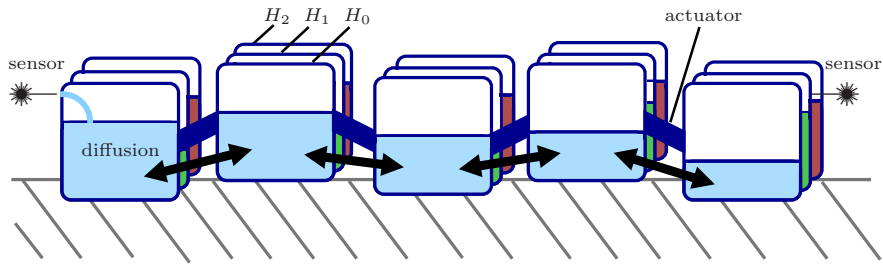}}
    \put(0,22){\large \Laserbeam}
    \put(110,22){\large \reflectbox{\Laserbeam}}
    \put(0,27){\scriptsize sensor}
    \put(90,33.5){\scriptsize actuator}
    \put(35,33.5){\scriptsize $H_2$}
    \put(40.5,33.5){\scriptsize $H_1$}
    \put(46,33.5){\scriptsize $H_0$}
    \put(10,15){\scriptsize diffusion}
    \put(112,27){\scriptsize sensor}
  \end{picture}
  \caption{\label{fig:ahhsPicto}Sketch of the hormone dynamics and
    diffusion processes in an organism. Each module holds different
    hormones with different concentrations, hormones diffuse through
    the organism based on a diffusion coefficient evolved individually
    for each hormone, module locations (e.g., elevation) are not
    relevant for diffusion; sensor settings simplified, actually four
    proximity sensors per module.}
\end{figure}

\subsection{AHHS1}
We call the AHHS, initially presented in
\citep{schmickl10a,schmickl09a}, AHHS1. An AHHS consists of
a set of hormones and a set of rules. On the one hand, it defines
production/decay rates and diffusion coefficients for each hormone. On
the other hand, it defines by rules the production through sensors and
interaction of hormones as well as their influence on actuators. There
are four types of rules. Sensor rules define the production of hormone
through sensor input. Actuator rules define the control of actuators
through hormone concentrations. Hormone rules define the interaction
between hormones, that is, one hormone triggers the production of
another hormone (or itself). Additionally, there is an idle rule to
allow a direct deactivation of rules through mutations. Rules are
triggered at runtime, if a certain threshold is reached (sensor values
in case of sensor rules or hormone concentrations in case of hormone
rules). The amount of produced hormone or the actuator control value
are linearly depending on the controlling sensor or hormone
respectively (`$\lambda x+\kappa$'). For more details see
\citet{schmickl10a}.

\subsection{AHHS2}
Based on AHHS1 we designed an improved variant called AHHS2. The
guiding principle of this improved controller design was to gain
higher evolvability by creating smoother fitness landscapes. There
were three main changes.

First, we introduced an additional rule type that implements nonlinear
hormone-to-hormone interactions in the general form of $\Delta
x/\Delta t=xy$, where $x$ is the considered hormone concentration and
$y$ is the hormone concentration of the influencing hormone that
triggers the considered rule. The idea is to increase the intrinsic
dynamics (basically transient behavior before equilibria are reached)
of the hormone network even without significant sensor input.

Second, a rule is not just triggered by exceeding or falling below a
threshold but is linearly weighted within a trigger window (i.e., a
tent function with a maximum of 1, defined by a center and a width,
see eq.~\ref{eq:trigger} below).

Third, the mutation of rule types in the form of discrete switches
seemed to be too radical. This was overcome by introducing a concept
of weights for rule types. Now, each rule can operate as any rule type
at the same time. Each rule has a weight for each of the five rule
types summing up to one (see Fig.~\ref{fig:ruleTypeWeights}). The
influence of a rule type is proportional to its weight, for example,
the sensor-rule aspect of a rule with a weight of 0.1 will produce
only 10\% of the hormone it would produce, if its weight would be 1,
see $w_{\mathcal L}$ in eq.~\ref{eq:rule:linearhormone}
below. A~mutation will now only change two rule weights by reducing
one by~$w$ and adding~$w$ to the other weight. In a well adapted
controller we would expect that the weights of a rule are mainly
concentrated on one or at most two rule types. Other weight
distributions should be transitional only because specialization
allows for better optimization.

The mathematical closed-form of this concept using the example of a
linear hormone rule type is

\begin{equation}
\label{eq:rule:linearhormone}
\mathcal L(t) =
w_{\mathcal L}\theta(H_k(t))(\lambda H_k + \kappa),
\end{equation}

where $\mathcal L(t)$ is the hormone amount that is to be added to the
considered hormone at time~$t$, $w_{\mathcal L}$ is the weight of the
linear hormone rule (see Fig.~\ref{fig:ruleTypeWeights}), $k$ is the
index of the input hormone and $H_k$ is its concentration, $\lambda$
is the dependent dose, $\kappa$ the fixed dose. $\theta$ is called
trigger function and defined by

\begin{equation}
\label{eq:trigger}
\theta(x) = 
\begin{cases}
  \frac{1}{\eta}(\eta-|x-\zeta|) & \text{if } |x-\zeta|<\eta\\
  0 & \text{else}
\end{cases},
\end{equation}

for trigger window center~$\zeta$ and trigger window width~$\eta$. For
a more detailed introduction of AHHS2 and for a comparison of the AHHS
approach to the standard ANN approach, see \citet{hamann10b}.

\setlength{\unitlength}{0.006\textwidth}
\begin{figure}
  \centering
  \begin{picture}(120,65)
    \put(0,10){\includegraphics[angle=0,width=120\unitlength]
      {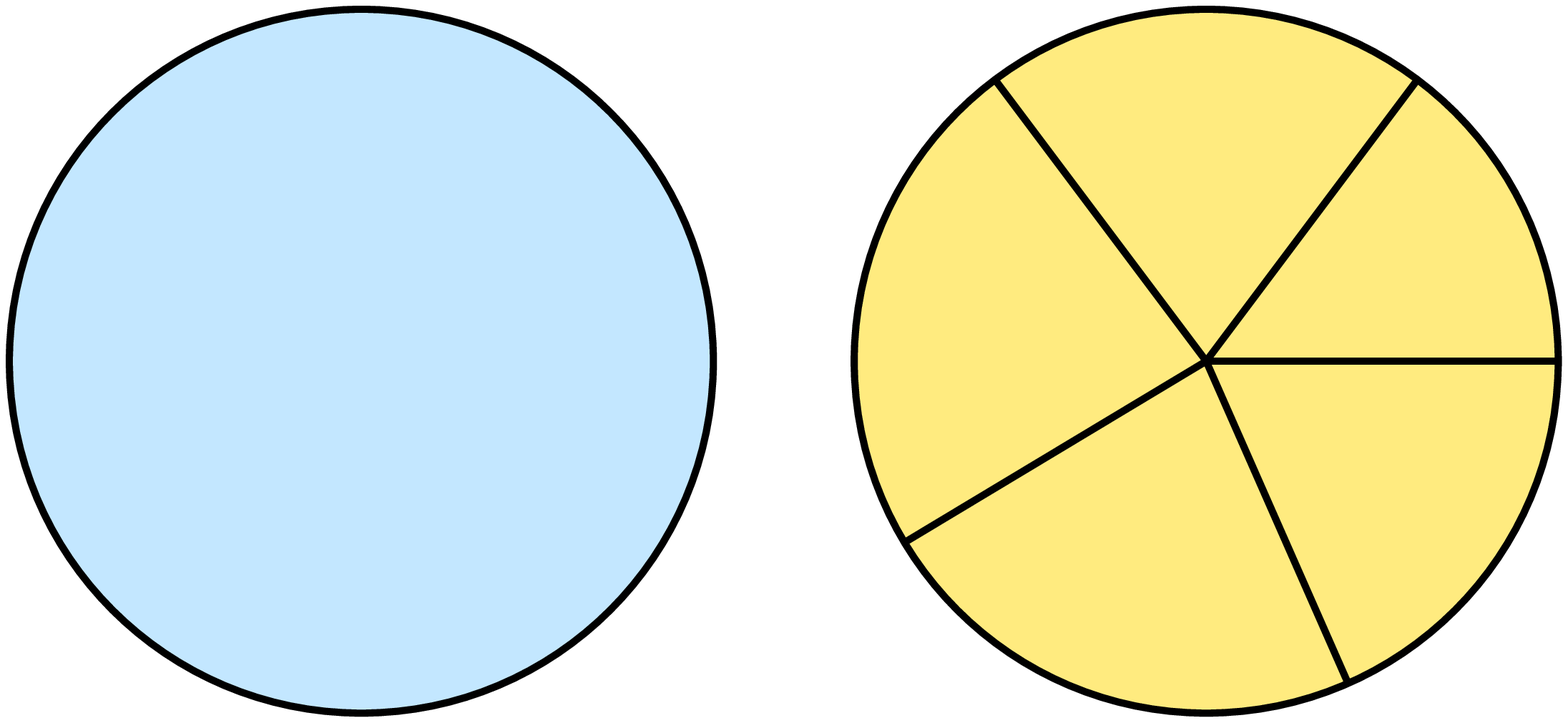}}
    \put(20,38){\footnotesize one of}
    \put(13,32){\footnotesize any rule type}
    \put(20,3){\footnotesize AHHS1}
    \put(85,3){\footnotesize AHHS2}
    \put(91,46){\begin{turn}{90}\scriptsize sensor r.\end{turn}}
    \put(98,39){\begin{turn}{40}\scriptsize l. horm. r.\end{turn}}
    \put(98,32){\begin{turn}{-40}\scriptsize nonl. h. r.\end{turn}}
    \put(82,14){\begin{turn}{65}\scriptsize actuator r.\end{turn}}
    \put(73,38){\begin{turn}{0}\scriptsize idle\end{turn}}
  \end{picture}
  \caption{\label{fig:ruleTypeWeights}Rule type weights of the AHHS2
    approach compared to AHHS1 (abbreviations: sensor rule, linear
    hormone rule, nonlinear hormone rule, actuator rule).}
\end{figure}

Note that the rule parameters (fixed dose, input hormone, trigger
window etc.) are correlated via the rule types. For example, the input
hormone is used for both the linear and the nonlinear hormone rule. If
we would allow independent parameters for each rule type the genome
(encoding of the controller) size would be increased by a factor of
about three. This is a tradeoff in the complexity of the genome and,
for example, a difficulty when analyzing the results. This is related
to the completeness-vs-compactness
challenge~\citep{mataric_1996_challenges}.

\section{Investigated scenarios}

Our main focus is on the field of modular robotics and our main
concern is whether we are able to evolve fast locomotion in the gait
learning task. Still, we tested the AHHS approach also in an inverted
pendulum task as well, due to its lower computational complexity.

\subsection{Inverted pendulum}

In addition to the gait learning task, we tested the AHHS approach in
a task that is easier to handle: balancing the inverted pendulum (see
Fig.~\ref{fig:pendulum}). The computational demand of the gait
learning task is very high due to the sophisticated simulation of
physics. We satisfy the need for a simulation of lower computational
complexity by introducing the inverted pendulum task. Higher
statistical significance of the results can be reached within
reasonable time of computation. The original inverted pendulum is only
slightly related to a real robotic task. Therefore, we adapted it to
our requirements. The sensors are noisy (equally distributed and
uncorrelated in time, $\pm 2.3\%$) and sampling rates of sensors are
low which is documented by the relation between the cycle
length~$\tau$ and the maximal angular velocity of $0.05\pi [1/\tau] =
9^\circ[1/\tau]$. The pendulum can move up to $9^\circ$ between two
calls of the controller. The controller has little time to adapt to
new configurations. Furthermore, the sensors do not deliver actual
angles and positions directly but partitioned onto several sensors and
also relative rather than absolute (distance to wall instead of the
crab's position etc.). The AHHS controls two outputs, left
actuator~$A_0$ and right actuator~$A_1$, while the speed control of
the crab is determined by their difference. The pendulum is started in
the lower equilibrium position, so the nonlinear up-swinging phase is
included. Combined with the sensor noise it is impossible for the
controller to balance the pendulum in the upper equilibrium
position. So the task stays dynamic and the controller is exposed to
new situations constantly. The fitness function is the summation over
all time steps of the angular distance to the top position in radians.

\setlength{\unitlength}{0.0034\textwidth}
\begin{figure}
  \centering
    \begin{picture}(130,75)
      \put(0,0){\includegraphics[angle=0,width=130\unitlength]
        {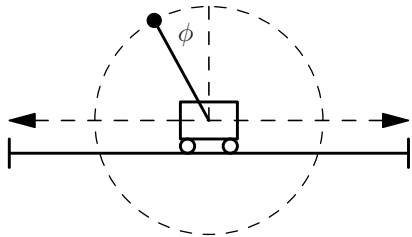}}
      \put(55,63){$\phi$}
    \end{picture}
    \caption{\label{fig:pendulum}Inverted pendulum, pendulum free to
      move full $360^{\circ}$ mounted on the crab that moves in one dimension
      (left/right) bounded by walls.}
\end{figure}

\subsection{Gait learning in multi-modular robotics}

Gait learning in legged robotics is a commonly studied task in
evolutionary robotics as reported by
\citet{nelson_2009_fitness}. However, here we investigate
gait learning in multi-modular robotics. Each module consists of one
hinge and we connect five modules. These five hinges are controlled
decentrally although the modules have a low-level communication
channel by means of diffusing hormones.

In contrast to the standard tasks of gait learning and collision
avoidance, the challenge of gait learning in multi-modular robotics is
more complex. The resulting gait is emergent due to the decentral and
cooperative control of the actuators. In addition, there are several
conceptionally different solutions, that is, different techniques of
locomotion with good performance (e.g., caterpillar-like, erected
walk, small jumps).

In each module the same controller is executed. Therefore, the gait
learning task includes several sub-tasks. The organism has to break
the symmetry (head and tail), synchronize through collective
cooperation, and start moving into a common direction. This
synchronization aspect is similar to the gait learning task for a
legged robot with HyperNEAT by \citet{clune09}.

\setlength{\unitlength}{0.006\textwidth}
\begin{figure}
  \centering
  \begin{picture}(120,70)
    \put(0,0){\includegraphics[angle=0,width=120\unitlength]
      {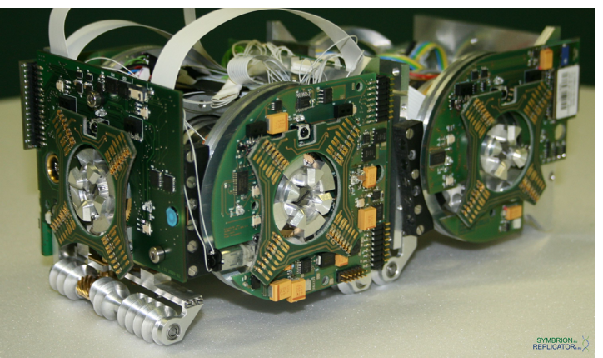}}
  \end{picture}
  \caption{\label{fig:prototype}Two connected prototypes of the projects \citet{symbrion} and \citet{replicator}.}
\end{figure}

All of this work is based on simulations as the actual hardware is not
yet available (see Fig.~\ref{fig:prototype} for a current prototype of
Symbrion and Replicator~\citep{symbrion,replicator}). We use the
simulation environment Symbricator3D by~\citet{winkler09} that was
developed for these projects. We use the current prototype design in
the simulation (imported CAD data) as described
in~\citep{levi10}. However, we simplified the sensor setting to four
proximity sensors (equally distributed around the robot shifted by
90~degrees: upwards, forwards, downwards, backwards). Symbricator3D is
based on the game engine Delta-3D and currently uses the Open Dynamics
Engine for the simulation of dynamics. The simulation of friction and
momentum is important because the evolved gait behaviors rely on
them. A drawback is that high computational complexity limits the
number of evaluations in our evolutionary runs. We are interested in
systems that evolve useful behaviors within a few hundred generations
and with small populations (order of 10).

We have tested the AHHS controllers with two variants of the
simulation framework. In the first version, the forces in the joints,
that connect the modules, were damped and small displacements of the
modules at the joints were allowed (i.e., simulation reacts moderately
to big forces). It turned out that caterpillar-like locomotion was
favored because the damped joints support wave motion. In the second
version, the joints were fully fixed. In this version of the
simulation the evolution of locomotion is more difficult which will be
reflected by the best fitnesses in the following.

We start the scenario with five robot modules which are simply
connected in a chain. Initially this robotic organism is placed in the
center of the arena. In order to increase the complexity of the gait
learning task, the central area is surrounded by a low wall forming a
square (its height is about half the height of a robot
module). Outside the wall several cubes are placed that could only be
sidestepped by the organism. An identical robot controller is uploaded
to the memory of all five modules. The robot modules have to figure
out their position (their role within the configuration), that is,
they have to break the symmetry of the configuration in order to
generate a coordinated gait. This is, for example, possible because of
different outputs of proximity sensors depending on the modules'
positions. There are three classes of modules defined by their
characteristic sensor inputs: front module, back module, and modules
in between. We use identical controllers because we want to apply them
to dynamic body shapes in our future work and also a single module
should have all functionality. Hence, uploading heterogeneous
controllers with predefined roles would not be an option. In addition,
using self-organized role assignment will allow for high scalability
(using the same controller for different body sizes), plasticity
(reorganization of roles in changing body shapes), and new role types
might emerge that were unthought of by the human designer.

The fitness is defined by the covered distance of the organism. It is
an aggregate fitness function~\citep{nelson_2009_fitness} that
evaluates the organism's performance as a whole. Although the
organisms might achieve advancements early in the evolutionary run,
there is a bootstrapping problem. For example, the downward proximity
sensors will not give significant input until the organism has figured
out how to erect the modules in the middle. In addition, controllers
cannot evolve special techniques to climb the wall before they have
actually managed to move the organism there to explore it.

\section{Results and discussion}

\subsection{Inverted pendulum}

The evolutionary runs of the inverted pendulum were performed with a
population of 200~randomly initialized controllers. The AHHS was set
to 15~hormones. For AHHS1 60 rules were used and 15 for AHHS2. The
runs were stopped after 200~generations. Linear proportional selection
was used and elitism was set to one. The mutation rate was 0.15 per
gene with a maximal, absolute change of range~0.1. The recombination
(two-point crossover) rate was~0.05.

For this task we configured AHHS with a left and a right compartment. The
left compartment incorporates the left actuator~$A_0$, the left proximity
sensor, the sensors giving the angles of the pendulum when it is in the
left half etc. and for the right compartment respectively.

The comparison of the best controllers of each run is shown in
Fig.~\ref{fig:pendCompare:fit}. In this scenario, AHHS2 performs
significantly better than AHHS1 although in terms of evolution speed
there is no significant difference (see
Fig.~\ref{fig:pendCompare:gen}). The AHHS2 design is the better choice
in this task. The cause of the advantage of AHHS2 over AHHS1 in this
task compared to the indistinct situation in the gait learning task is
unclear. In future studies we will investigate whether this trend will
also be observed in more complex tasks from the domain of
multi-modular, evolutionary robotics.

\setlength{\unitlength}{0.0036\textwidth}
\begin{figure}
  \centering
  \subfigure[\label{fig:pendCompare:fit}fitness (Wilcoxon $p<0.05$)]{
    \begin{picture}(120,75)
    \put(5,89){\includegraphics[angle=270,width=120\unitlength]
        {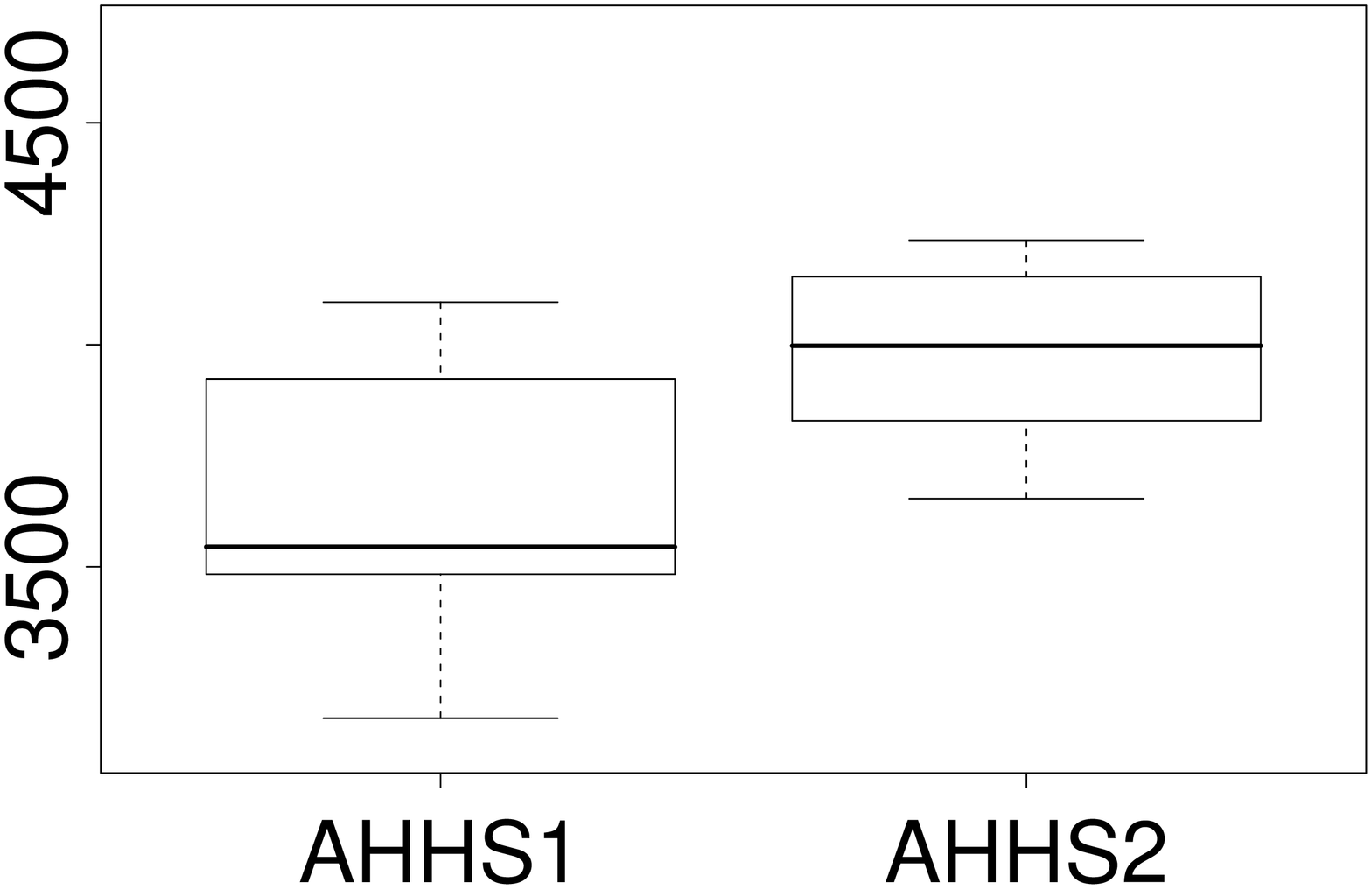}}
      \put(34,-1){\tiny $n=8$}
      \put(79,-1){\tiny $n=10$}
      \put(44,63){\line(1,0){47}}
      \put(64,66){$\ast$}
      \put(-1,35){\begin{sideways}\small fitness\end{sideways}}
    \end{picture}
  }
  \subfigure[\label{fig:pendCompare:gen}generation]{
    \begin{picture}(110,75)
    \put(5,89){\includegraphics[angle=270,width=120\unitlength]
        {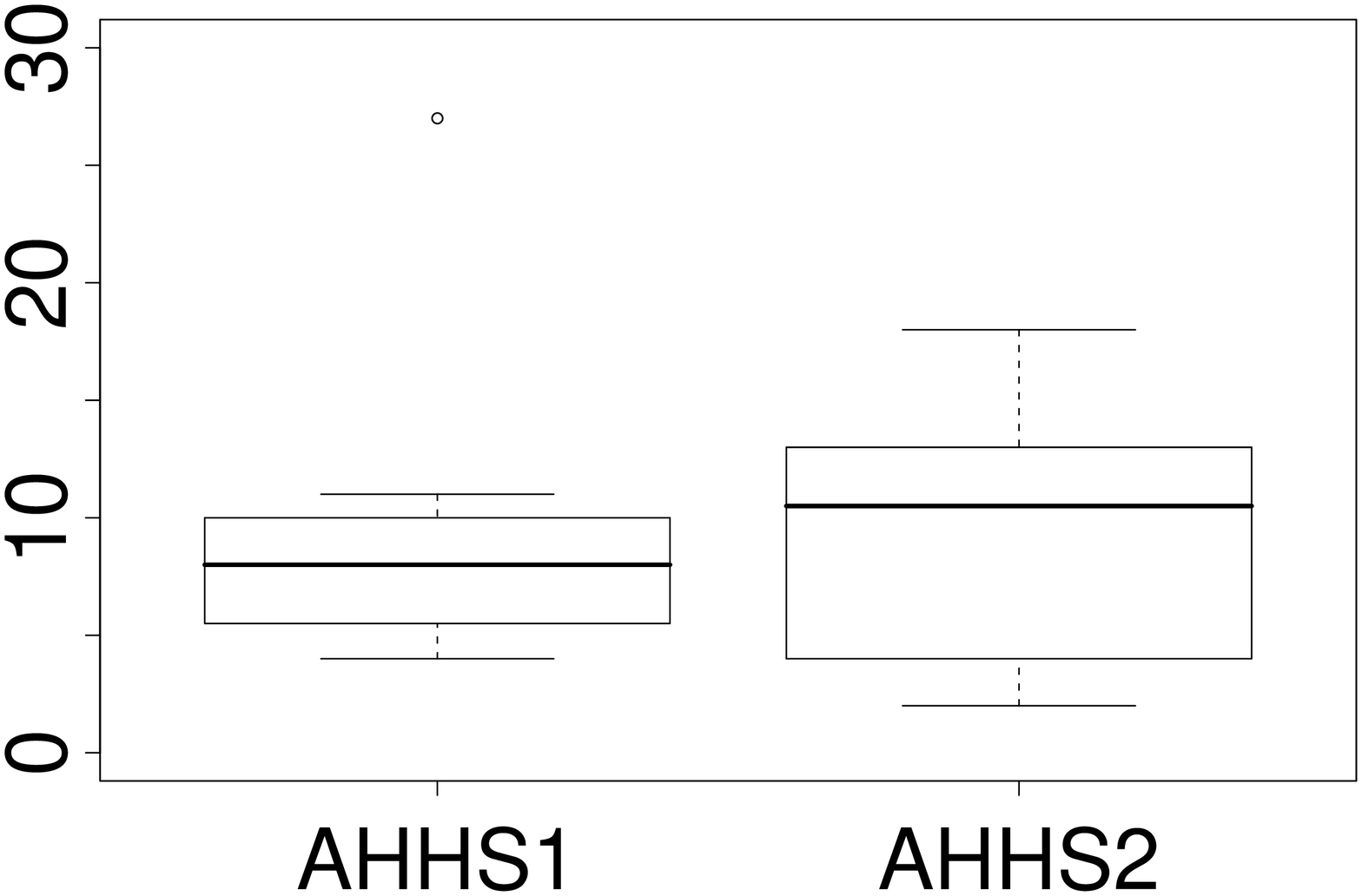}}
      \put(34,-1){\tiny $n=8$}
      \put(79,-1){\tiny $n=10$}
      \put(-2,24){\begin{sideways}\small generation\end{sideways}}
    \end{picture}
  }
  \caption{\label{fig:pendCompare}Inverted pendulum, AHHS1 with 60
    rules, AHHS2 with 15 rules, comparison of fitness and evolution
    speed (generation when 75\% of max. fitness was reached).}
\end{figure}

One of the best evolved AHHS2 controllers showing interesting behavior
is analyzed in the following\footnote{\scriptsize
  \url{http://heikohamann.de/pub/hamannEtAlAlife2010pend.mpg}}. While
it is not possible to keep the pendulum in the upper equilibrium for
longer time due to noise, the controller still tries to maximize the
time the pendulum is close to the upper equilibrium mostly by small
displacements of the crab. The controller is mainly based on one
hormone ($H_0$), and four rules (see
Fig.~\ref{fig:bestIndAnaPend:rules}). Sensor~$S_0$ reaches its
maximum, if the pendulum approaches~$\phi=0$ (top position) from the
left. It triggers small displacements of the crab to the right,
a~behavior that keeps the pendulum turning counterclockwise with slow
passes at the top position. Sensor~$S_9$ gives the intensity of
negative angular velocities of the pendulum (clockwise turns) and
triggers moves of the crab to the left. The proximity sensors are not
used at all. The walls are avoided by the crab movements depending on
position and turning direction of the pendulum. Hence, the position of
the crab is virtually encoded in the motion of the pendulum.

\setlength{\unitlength}{0.004\textwidth}
\begin{figure}
  \centering
  \begin{picture}(120,120)
    \put(0,0){\includegraphics[angle=0,width=120\unitlength]
      {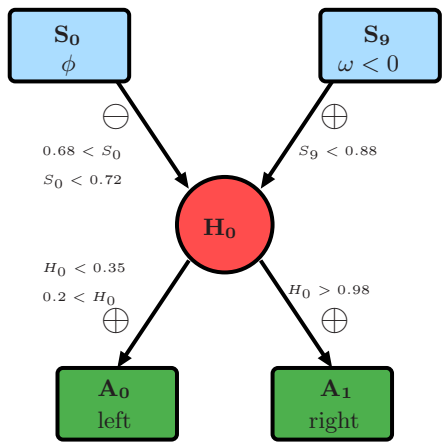}}
    \put(13,110){\small ${\bf S_0}$}
    \put(15,102){\small $\phi$}


    \put(98,110){\small ${\bf S_9}$}
    \put(91,102){\small $\omega<0$}
    \put(54,57){\small ${\bf H_0}$}
    \put(25,13){\small ${\bf A_0}$}
    \put(25,4){\small left}
    \put(86,13){\small ${\bf A_1}$}
    \put(83,4){\small right}
    \put(26,87){\Large $\ominus$}
    \put(10,79){\tiny $0.68<S_0$}
    \put(10,71){\tiny $S_0<0.72$}


    \put(86,87){\Large $\oplus$}
    \put(80,79){\tiny $S_9<0.88$}

    \put(26,31){\Large $\oplus$}
    \put(77,41){\tiny $H_0>0.98$}
    \put(86,31){\Large $\oplus$}
    \put(10,47){\tiny $H_0<0.35$}
    \put(10,39){\tiny $0.2<H_0$}
  \end{picture}
  \caption{\label{fig:bestIndAnaPend:rules}Inverted pendulum, analysis
    of one of the best evolved AHHS2 controllers; only most relevant
    rules of the evolved behavior are shown.}
\end{figure}

See Fig.~\ref{fig:bestIndAnaPend:dynamics} for the sensor, hormone,
and actuator dynamics. This sample run begins with an initial ($t<50$)
move of the crab from the center to the outer left due to transient
dynamics of $H_0$ in the left compartment (see
Fig.~\ref{fig:bestIndAnaPend:hormone}). This motion implements the
up-swinging of the pendulum and is followed by ten small displacements
of the crab to the right to keep the pendulum swinging
counterclockwise. At $t=1093$ the turning direction of the pendulum
changes (see Fig.~\ref{fig:bestIndAnaPend:sa}). A~sequence of
right-left movements is initiated to reestablish the counterclockwise
turning. Later at $t=1933$ a phase of low angular velocity is reached
which causes irregular movements of the crab that hold the pendulum
close to the top position.

\setlength{\unitlength}{0.008\textwidth}
\begin{figure}
  \centering
  \subfigure[\label{fig:bestIndAnaPend:hormone}most relevant hormone $H_0$ (upper
    and lower half, red), actuator left $A_0$ (upper half, black),
    right $A_1$ (lower half, black)]{
    \begin{picture}(120,27)
      \put(0,57){\includegraphics[angle=270,width=120\unitlength]
        {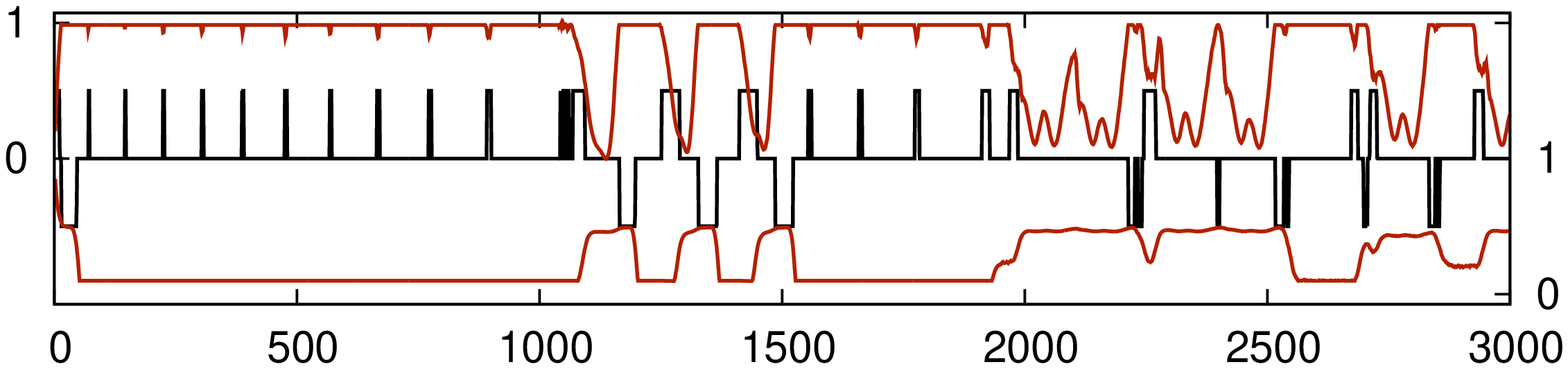}}
      \put(61,-2){$t$}
      \put(116,8){\begin{sideways}left\end{sideways}}
      \put(2,16){\begin{sideways}right\end{sideways}}
    \end{picture}
  }
  \subfigure[\label{fig:bestIndAnaPend:sa}pendulum angle sensor $S_0$ for
    $0<\phi<\pi/2$ (purple), negative angular velocity sensor $S_9$
    (lower half, yellow)]{
    \begin{picture}(120,27)
      \put(0,57){\includegraphics[angle=270,width=120\unitlength]
        {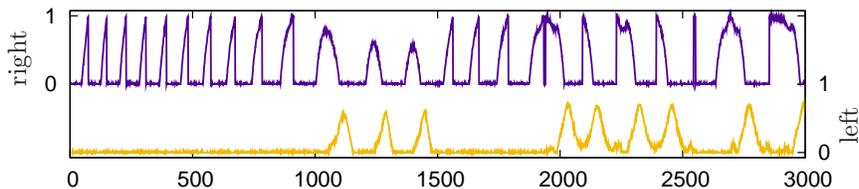}}
      \put(61,-2){$t$}
      \put(116,8){\begin{sideways}left\end{sideways}}
      \put(2,16){\begin{sideways}right\end{sideways}}
    \end{picture}
  }
  \caption{\label{fig:bestIndAnaPend:dynamics}Inverted pendulum, most
    relevant hormone, sensors, and both actuator control values for
    both compartments (left and right) of the evolved behavior.}
\end{figure}

\subsection{Gait learning}

The evolutionary runs of the gait learning task were performed with a
population of 20~randomly initialized controllers. The configuration
of the AHHS was set to 5~hormones. The number of rules was varied
between 20 and 300. The runs were stopped after
200~generations. Linear proportional selection was used and elitism
was set to one. The mutation rate was 0.15 per gene (rule or hormone,
with a maximal, absolute change of range~0.1). The recombination
(two-point crossover) rate was~0.05. One run of the evolution (full
200 generations) took about 28~hours of CPU time (on a single core of
a standard, up-to-date desktop PC).

In the first version of the simulation (damped joints), the evolved
behaviors reach high fitness values for all investigated settings of
the AHHS (see Fig.~\ref{fig:botCompare}). Directly approaching the
wall yields a fitness of about 0.7, getting one half of the modules
over the wall yields a fitness of 0.8, and a fitness of above 1 is
reached, if the wall is overcome. Typically the evolved behaviors rely
on two or three of the five provided hormones only and make use of
less than ten rules. However, a too low number of rules results in too
little exploration of the behavior space. Based on preliminary tests
we decided to use 30 rules for AHHS2. One AHHS2 rule is potentially
active for each rule type, which corresponds to four active AHHS1
rules. However, AHHS2 cannot optimize the parameters for each rule
type individually. Still, we tested the AHHS1 with 120 rules and also
with a much higher number of 300 rules. The results show no
statistical significant differences but show in a trend that the AHHS1
does not reach comparable results as AHHS2 with corresponding rule
numbers. In addition, the behaviors evolved by AHHS1 show high
variance depending on the deterministic chaos through the complex
system (simulation of physics).

\setlength{\unitlength}{0.0036\textwidth}
\begin{figure}
  \centering
  \subfigure[\label{fig:botCompare:fit}fitness]{
    \begin{picture}(120,81)
    \put(5,95){\includegraphics[angle=270,width=120\unitlength]
        {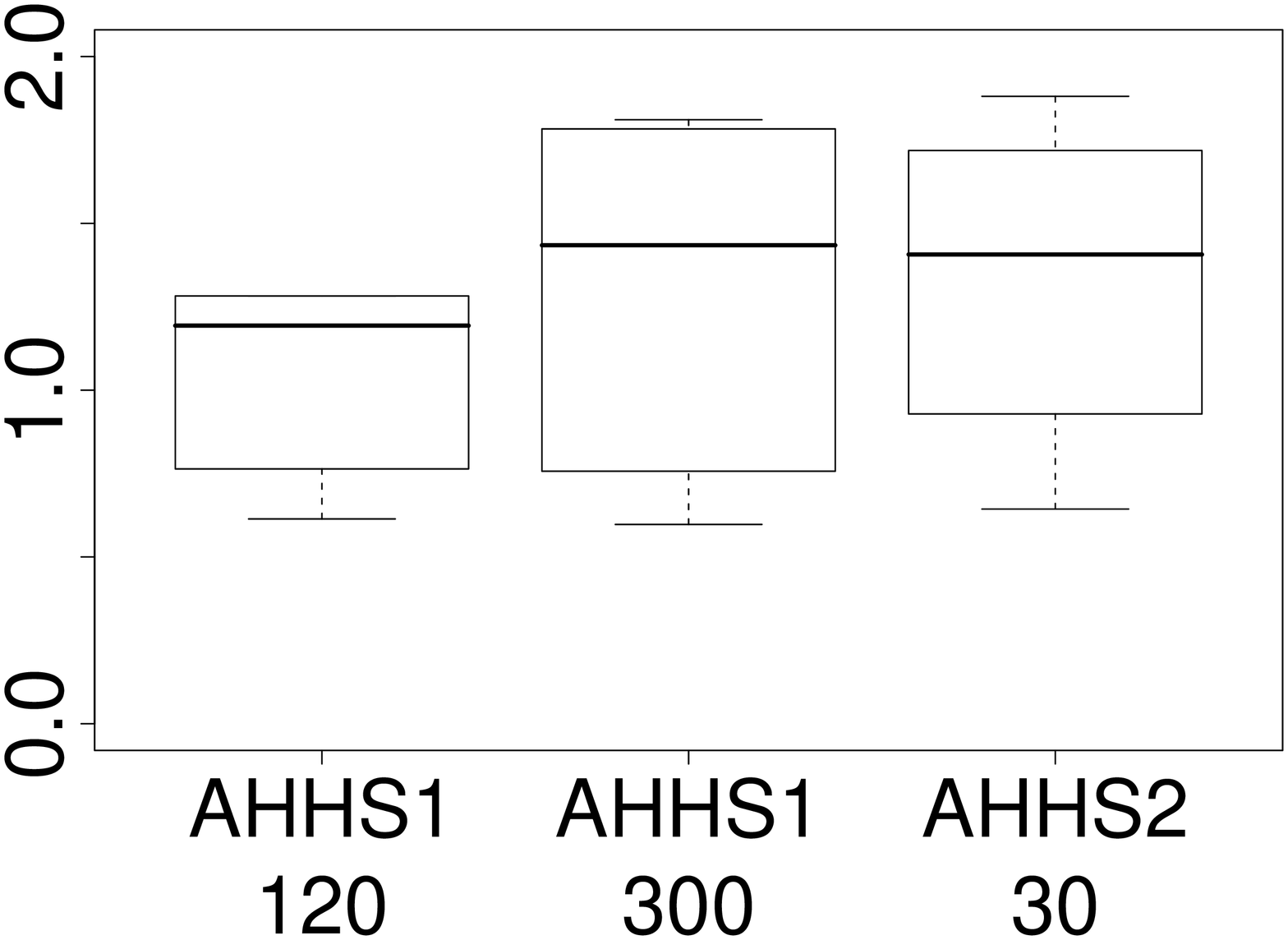}}
      \put(27,-1){\tiny $n=8$}
      \put(57,-1){\tiny $n=13$}
      \put(88,-1){\tiny $n=13$}
      \put(-2,40){\begin{sideways}\small fitness\end{sideways}}
    \end{picture}
  }
  \subfigure[\label{fig:botCompare:gen}generation]{
    \begin{picture}(120,80)
    \put(5,95){\includegraphics[angle=270,width=120\unitlength]
        {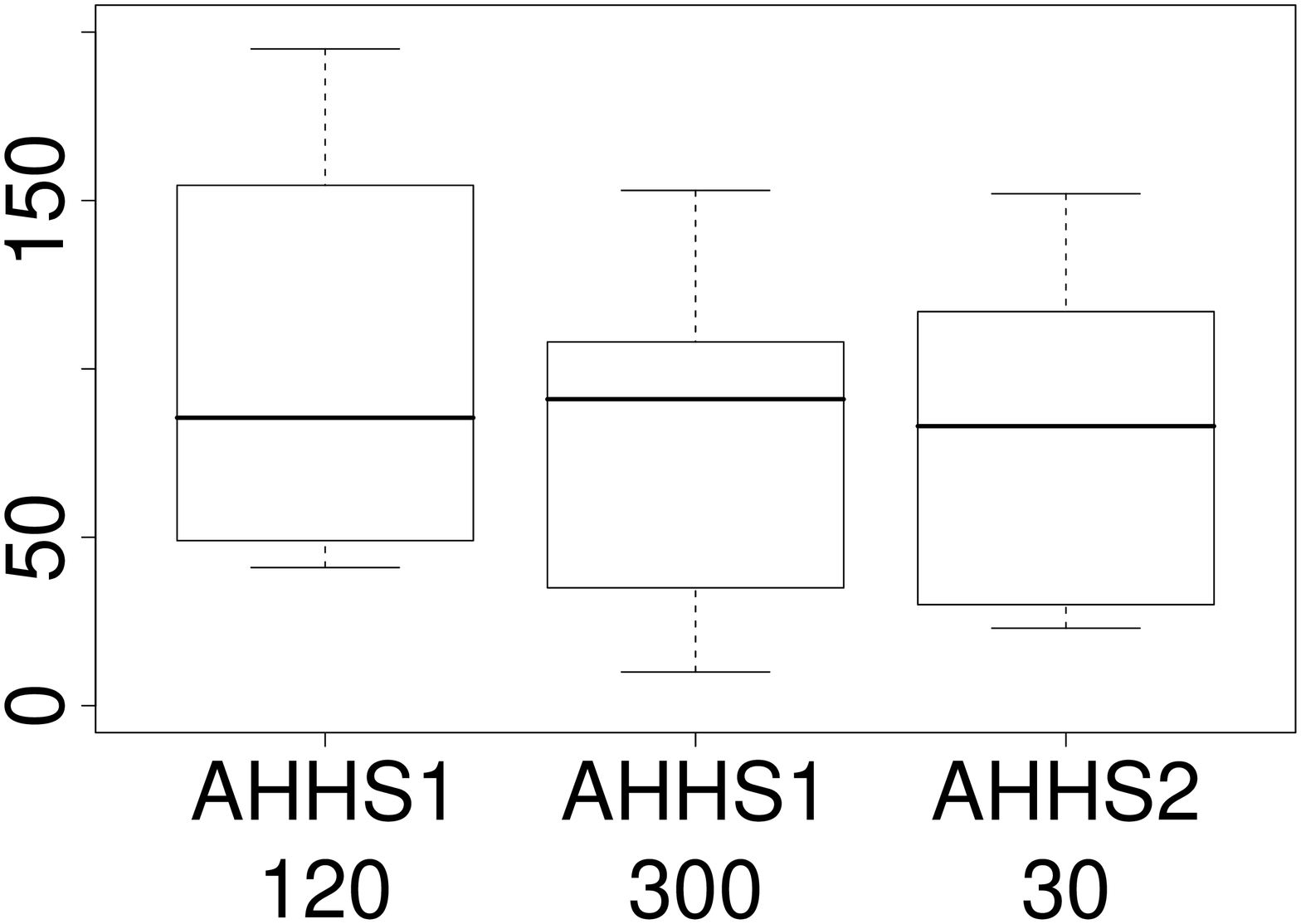}}
      \put(27,-1){\tiny $n=4$}
      \put(57,-1){\tiny $n=9$}
      \put(88,-1){\tiny $n=10$}
      \put(-2,31){\begin{sideways}\small generation\end{sideways}}
    \end{picture}
  }
  \caption{\label{fig:botCompare}5-module gait learning with damped
    joints, comparison of fitness and evolution speed, which is
    indicated by the generation in which 75\% of the overall
    max. fitness ($1.41=0.75\times 1.88$) was reached (if at all).}
\end{figure}


Using the second version of the simulation (fixed joints), we have
tested smaller differences in the number of rules between AHHS1 and
AHHS2. The results show that the more realistic simulation of the
joints complicates the evolution of fast locomotion. However, the
favoring of caterpillar-like locomotion is reduced significantly and
especially in case of AHHS2 an unexpected vast
diversity\footnote{\scriptsize
  \url{http://heikohamann.de/pub/hamannEtAlAlife2010.mpg}} of
different locomotion paradigms is observed (see
Fig.~\ref{fig:diversity} for a short collection). Basically we
observed three classes of locomotion: erected walking behavior,
caterpillar-like locomotion, and locomotion through jumps. The
behaviors evolved using AHHS1 were less diverse. Quantifying these
differences will be the focus of future studies.

\setlength{\unitlength}{0.003\textwidth}
\begin{figure}[h]
  \centering
  \subfigure[\label{fig:diversity:1}walking]{
    \begin{picture}(120,81)
    \put(0,0){\includegraphics[angle=0,width=120\unitlength]
        {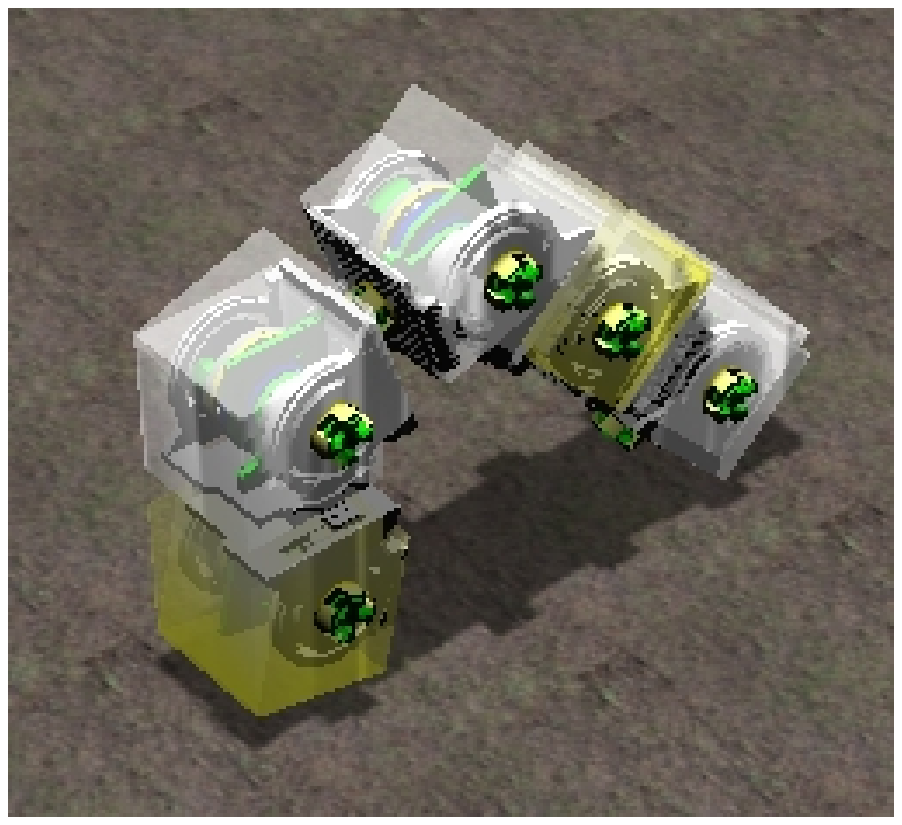}}
    \end{picture}
  }
  \subfigure[\label{fig:diversity:2}upside down over wall]{
    \begin{picture}(120,110)
    \put(0,0){\includegraphics[angle=0,width=120\unitlength]
        {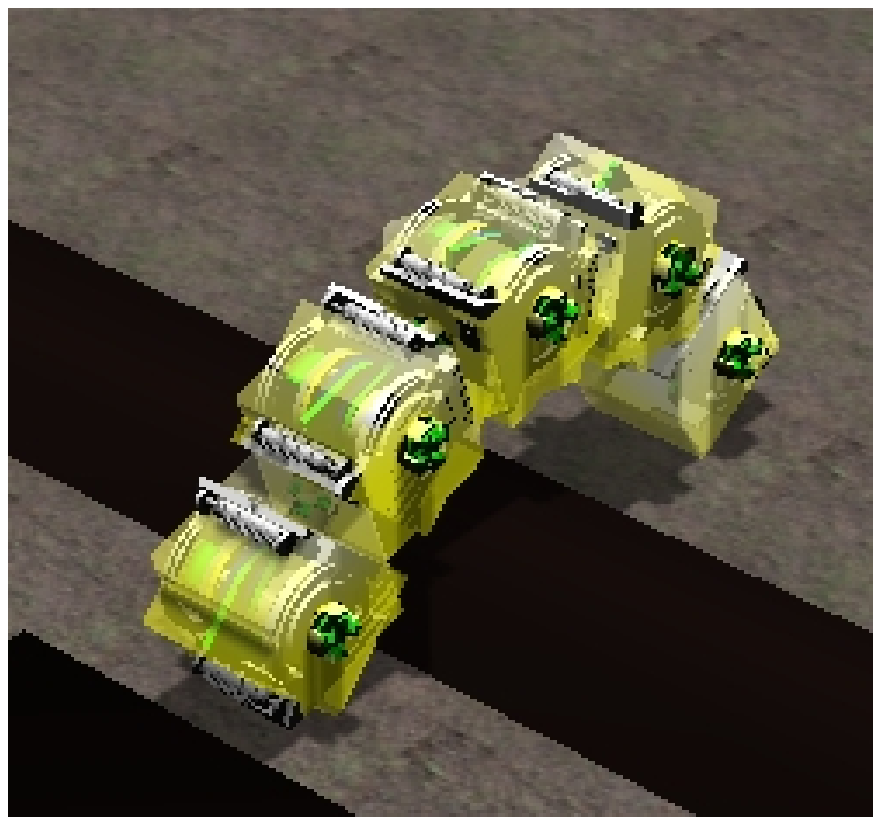}}
    \end{picture}
  }
  \subfigure[\label{fig:diversity:3}independent hinges]{
    \begin{picture}(120,88)
    \put(0,0){\includegraphics[angle=0,width=120\unitlength]
        {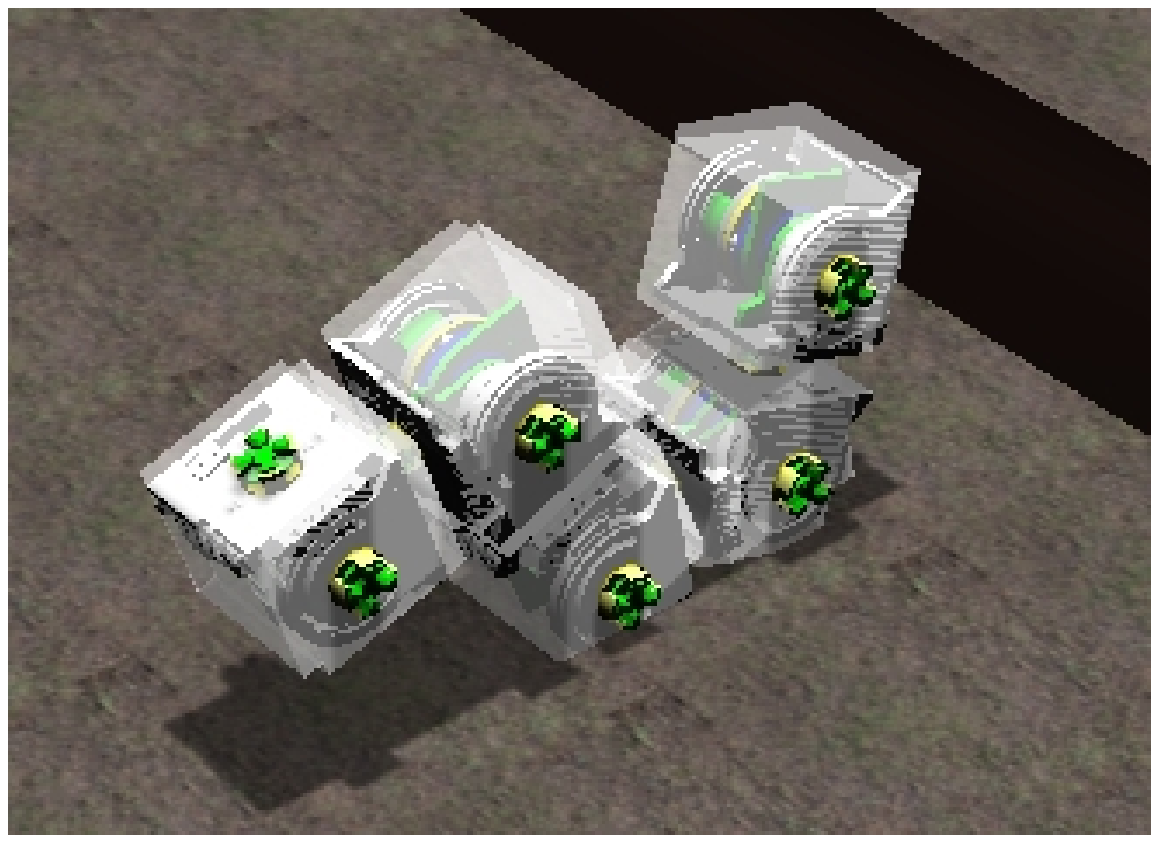}}
    \end{picture}
  }
  \subfigure[\label{fig:diversity:4}caterpillar-like]{
    \begin{picture}(120,85)
    \put(0,0){\includegraphics[angle=0,width=120\unitlength]
        {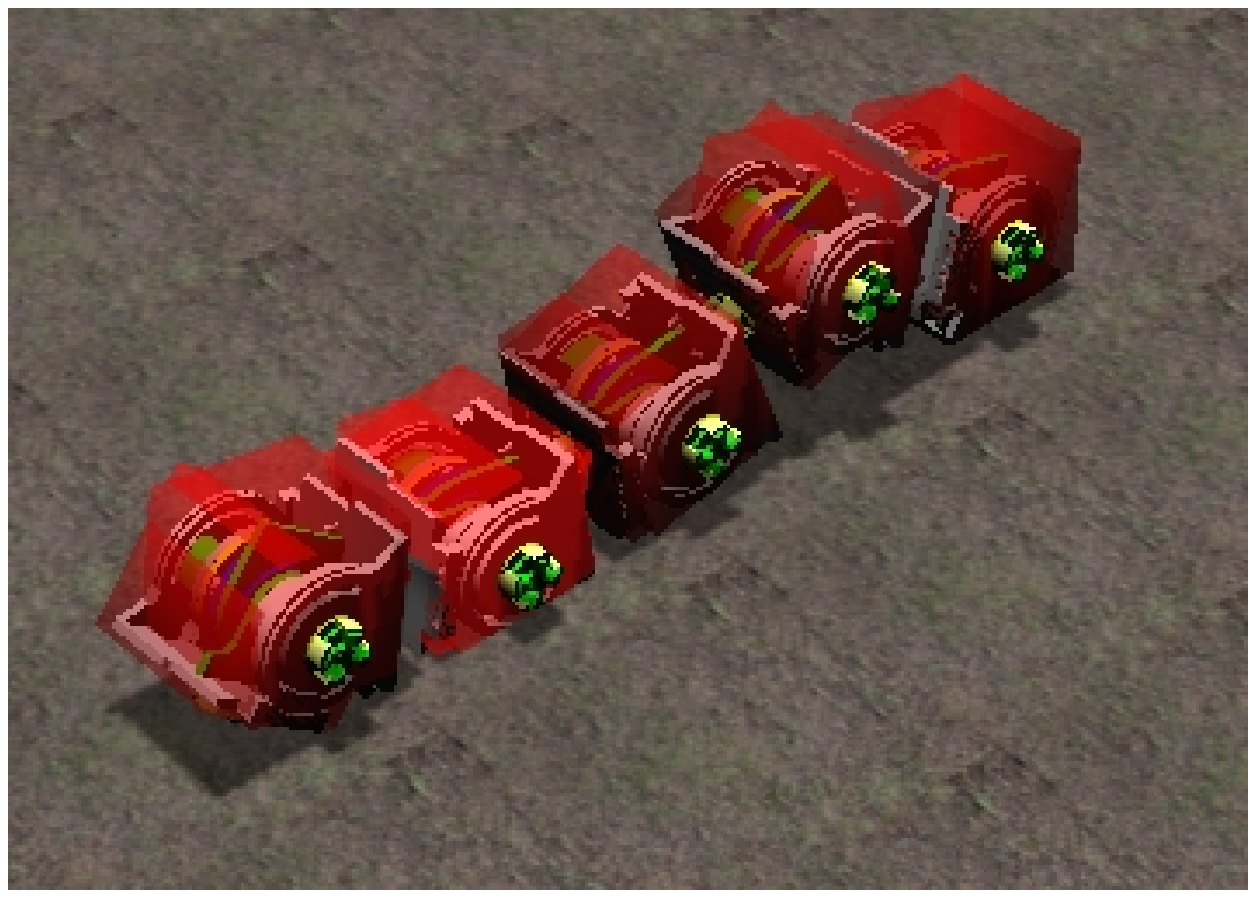}}
    \end{picture}
  }
  \subfigure[\label{fig:diversity:5}jumping]{
    \begin{picture}(120,130)
    \put(0,0){\includegraphics[angle=0,width=120\unitlength]
        {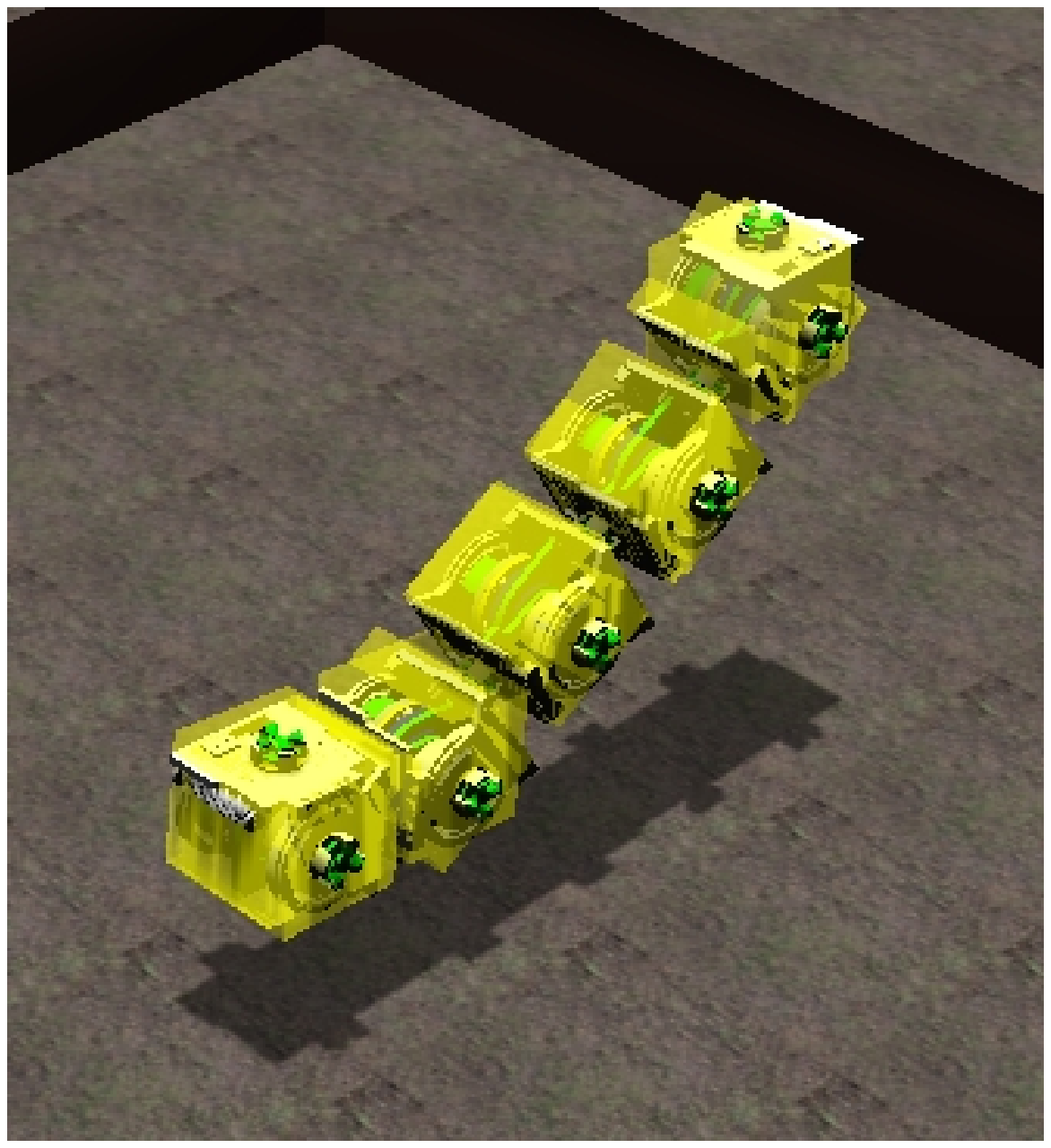}}
    \end{picture}
  }
  \subfigure[\label{fig:diversity:6}warping over the wall]{
    \begin{picture}(120,130)
    \put(0,0){\includegraphics[angle=0,width=120\unitlength]
        {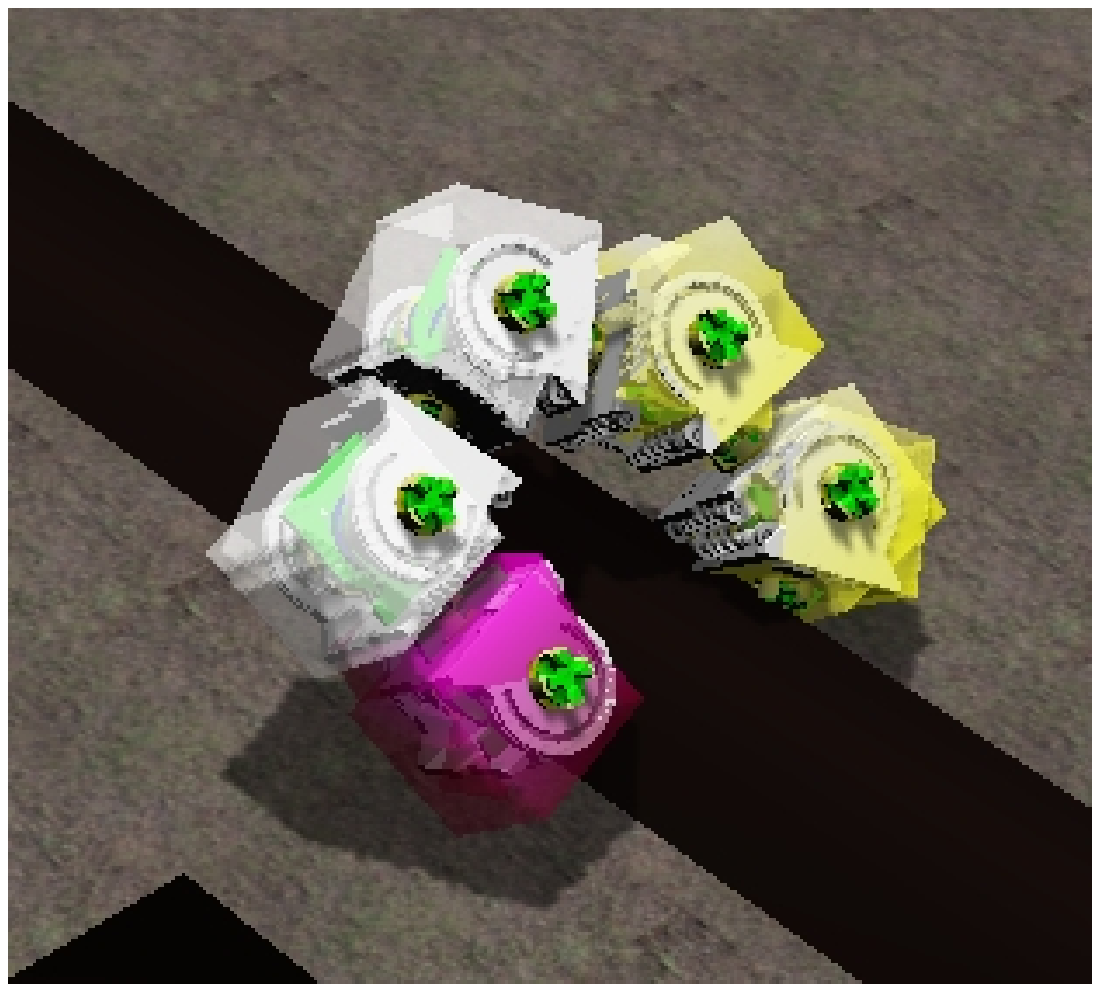}}
    \end{picture}
  }
  \caption{\label{fig:diversity}Screenshots showing the diversity of
    evolved locomotion paradigms (colors represent three selected
    hormones in the primary colors according to the RGB color model).}
\end{figure}

The comparison of the best evolved behaviors is shown in
Fig.~\ref{fig:ns_botCompare:fit} and the speed of evolution is shown
in Fig.~\ref{fig:ns_botCompare:gen}. 55\% of the AHHS2-runs with 50
rules and 38\% of the AHHS1-runs with 80 rules reach a best fitness
that is within 80\% of the theoretical maximum fitness of about
1.7. Significant results are only reached for AHHS1 with 20 rules
compared to both AHHS1 with 80 rules and to AHHS2 with 50
rules. Noticeable is the bad performance of AHHS2 with just 20 rules
both in terms of final best fitness and speed of evolution. From our
observations we speculate that the initial exploration (during few of
the early generations) of the search space (basically the
sensory-motor configurations) is a relevant feature. Identifying the
actual shortcoming of AHHS2 in this context is part of our future
research.

\setlength{\unitlength}{0.0036\textwidth}
\begin{figure}[t]
  \centering
  \subfigure[\label{fig:ns_botCompare:fit}fitness (Wilcoxon $p<0.05$)]{
    \begin{picture}(120,80)
    \put(5,95){\includegraphics[angle=270,width=120\unitlength]
        {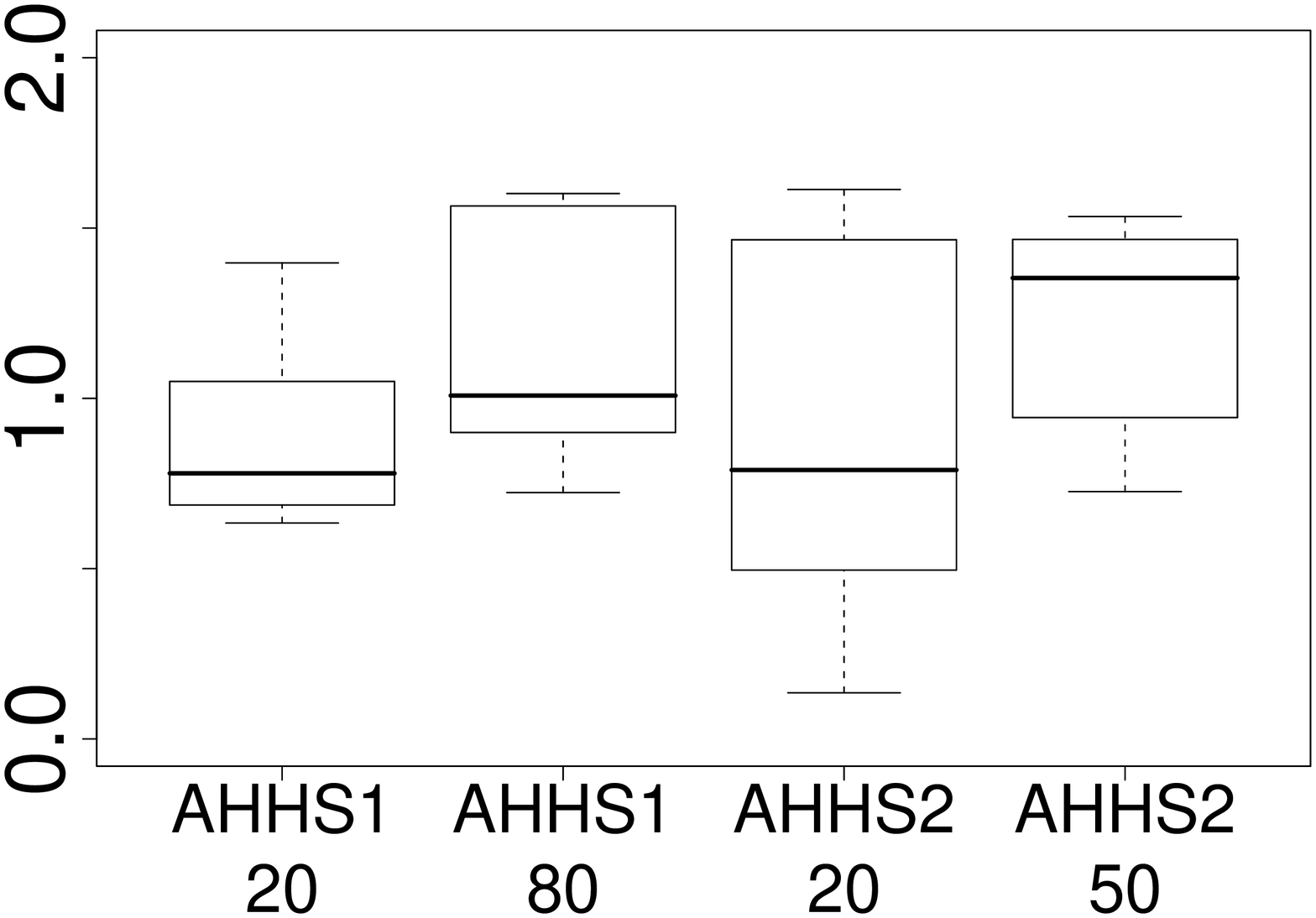}}
      \put(22,-1){\tiny $n=8$}
      \put(46,-1){\tiny $n=8$}
      \put(69,-1){\tiny $n=9$}
      \put(93,-1){\tiny $n=11$}
      \put(32,70){\line(1,0){25}}
      \put(43,71){\tiny $\ast$}
      \put(32,76){\line(1,0){70}}
      \put(65,77){\tiny $\ast$}
      \put(-2,40){\begin{sideways}\small fitness\end{sideways}}
    \end{picture}
  }
  \subfigure[\label{fig:ns_botCompare:gen}generation (Wilc. $p<0.05$)]{
    \begin{picture}(120,80)
    \put(5,95){\includegraphics[angle=270,width=120\unitlength]
        {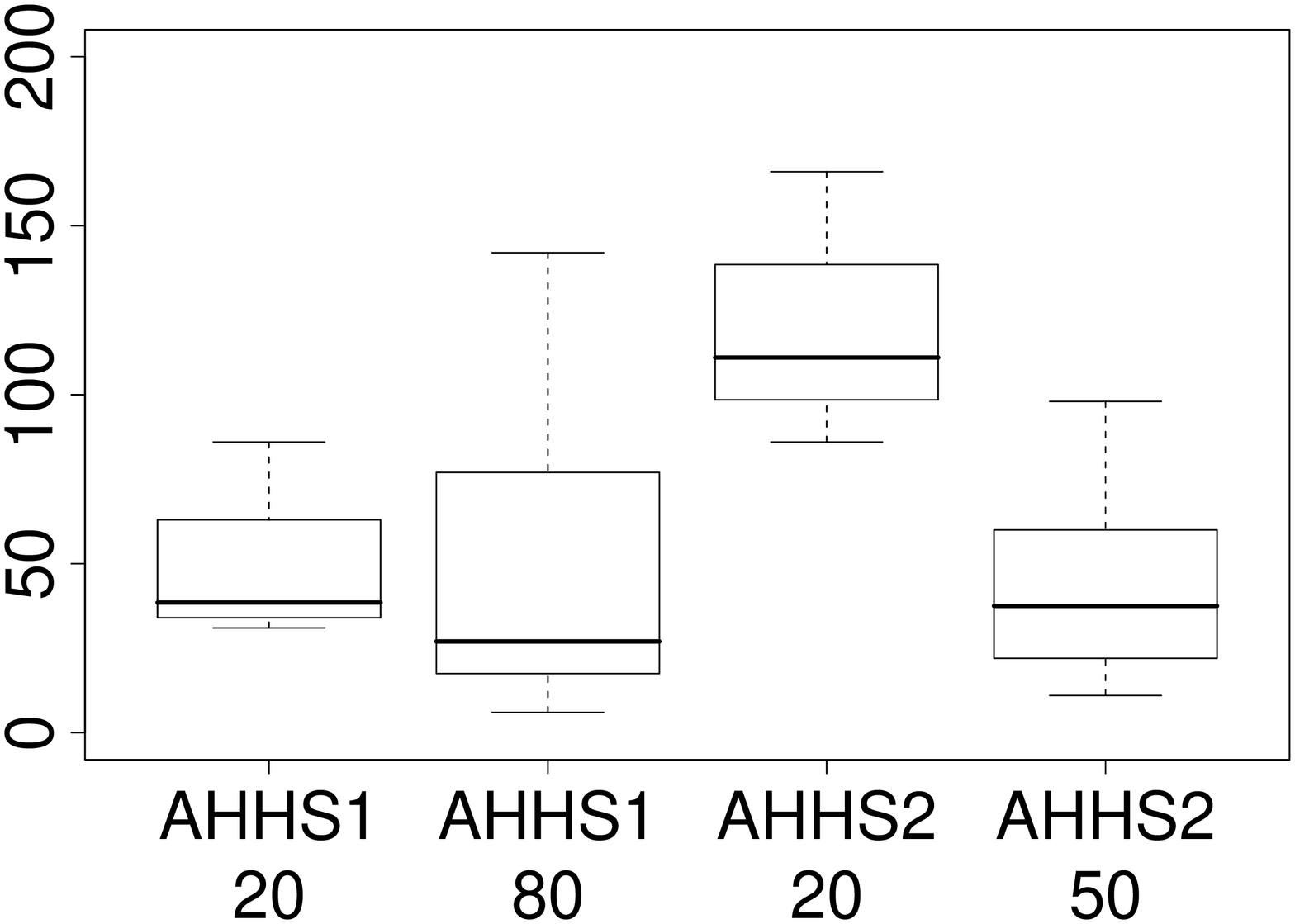}}
      \put(22,-1){\tiny $n=4$}
      \put(46,-1){\tiny $n=7$}
      \put(69,-1){\tiny $n=3$}
      \put(93,-1){\tiny $n=10$}
      \put(55,73){\line(1,0){24}}
      \put(65,74){\tiny $\ast$}
      \put(79,76){\line(1,0){23}}
      \put(89,77){\tiny $\ast$}
      \put(-2,31){\begin{sideways}\small generation\end{sideways}}
    \end{picture}
  }
  \caption{\label{fig:ns_botCompare}5-module gait learning with fixed
    joints, comparison of fitness and evolution speed, which is
    indicated by the generation in which 75\% of the overall
    max. fitness was reached (if at all).
  }
\end{figure}

One important aspect in the differences between the two controller
types seems to be the different triggering of rules in AHHS1 and
AHHS2. The behaviors of AHHS1 clearly show more fast-paced
movements. With damped joints this seems to be a disadvantage as
smooth movements are less likely. Using the fixed joints this
sometimes results in fast locomotion through little jumps.

The evolved structures are complex and the underlying processes are
often counter-intuitive. The in-depth analysis of individual behaviors
is alleviated by considering the number of steps a rule has been
active (triggered). Typically, about one third of the rules trigger
never or very seldom.

\subsection{Post-evaluation and analysis}

We have investigated the behavior of one of the best evolved AHHS2
controllers in the second version of the simulator. It shows a dynamic
caterpillar-like motion\footnote{\scriptsize\url{http://heikohamann.de/pub/hamannEtAlAlife2010ind.mpg}}. It
is noticeable that the rules show characteristics of specialization
and optimization. For example, often the (floating) index of the
output hormone is close to an integer (i.e., the rule's effect is
mostly limited to one hormone) and often rule weights are above 0.5
showing the specialization of those rules. For the investigated
controller we have identified three most relevant hormones: $H_2$,
$H_3$, and $H_4$. The angle of the hinge is mainly controlled by
hormones $H_3$ and $H_4$ (see Fig.~\ref{fig:bestIndAna:hsa}. High
values of $H_4$ turn the hinge towards $+90^\circ$ while any value of
$H_3>0$ turns the hinge towards $-90^\circ$. As a reinforcing effect
there is a hormone rule that decreases $H_4$, if $H_3>0$. $H_2$~shows
the influence by diffusion of hormones through the organism (see
Fig.~\ref{fig:bestIndAna:diff}. A~decreasing concentration in the back
module is consequently followed by a decrease in the second last,
middle, and second first module, hence, forming a hormone wave that is
propagating through the organism.
\setlength{\unitlength}{0.008\textwidth}
\begin{figure}
  \centering
  \subfigure[\label{fig:bestIndAna:hsa}Most relevant hormones $H_3$ (black) and $H_4$ (purple), and hinge control angle~$\phi$ (yellow).]{
    \begin{picture}(120,27)
    \put(0,57){\includegraphics[angle=270,width=120\unitlength]
        {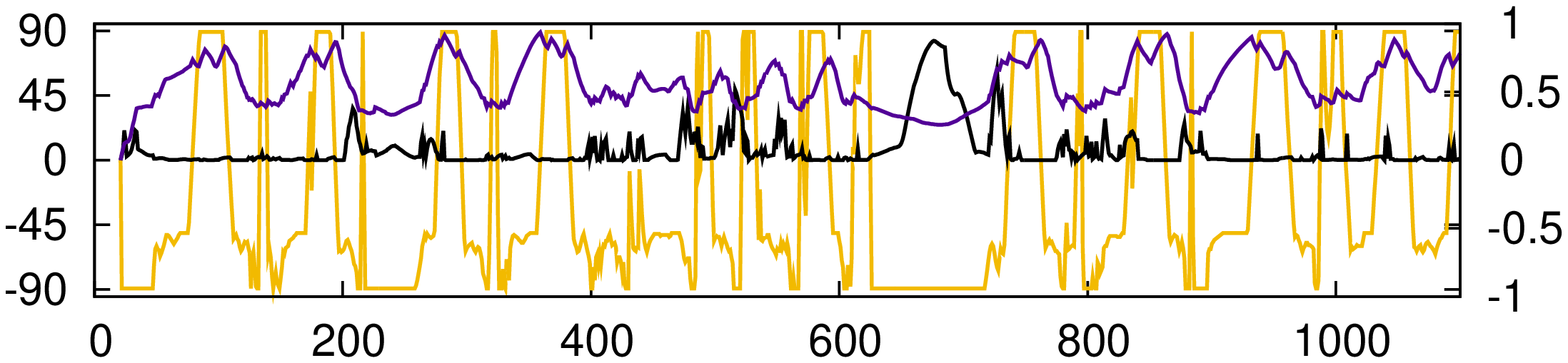}}
      \put(61,-2){$t$}
      \put(114,15){$H$}
      \put(1,15){$\phi$}
    \end{picture}
  }
  \subfigure[\label{fig:bestIndAna:diff}Hormone $H_2$ in all five modules,
    demonstrating the effect of diffusion (from front module to back: light to dark).]{
    \begin{picture}(120,27)
    \put(3,53){\includegraphics[angle=270,width=106\unitlength]
        {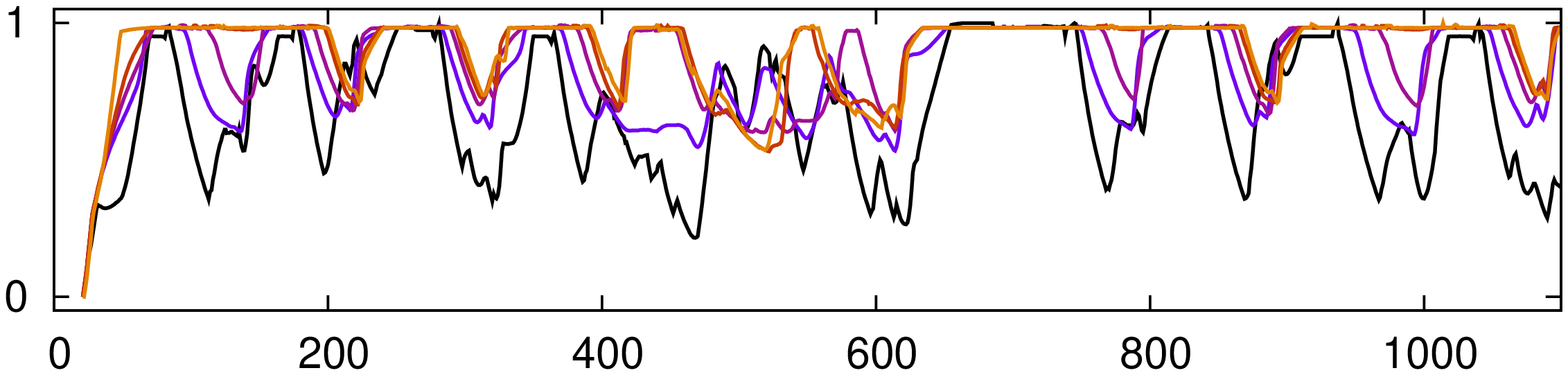}}
      \put(61,0){$t$}
      \put(1,15){$H$}
    \end{picture}
  }
  \caption{\label{fig:bestIndAna}5-module gait learning with fixed
    joints, analysis of the evolved behavior.}
\end{figure}
Finally, we investigated the influence of mutations. The leading
design paradigm of AHHS2 was to improve the causality of the mutation
operator (small changes in genome result in small changes in the
behavior). This was done exemplarily by taking an evolved controller
from each type. For both we produced 35~controllers by applying the
mutation operator once for each. The evaluated fitnesses of these
35~controllers are shown as a histogram in
Fig.~\ref{fig:fitnessAnalysis}. For AHHS1 the majority of mutated
controllers had a fitness of less than~0.2. For AHHS2 the majority of
mutated controllers reached about the original fitness. For both
types some controllers reached higher fitness due variance introduced
by deterministic chaos in the simulated physics.

\setlength{\unitlength}{0.0036\textwidth}
\begin{figure}
  \centering
  \subfigure[\label{fig:fitnessAnalysis:ahhs1}AHHS1]{
    \begin{picture}(120,85)
      \put(10,95){\includegraphics[angle=270,width=120\unitlength]
        {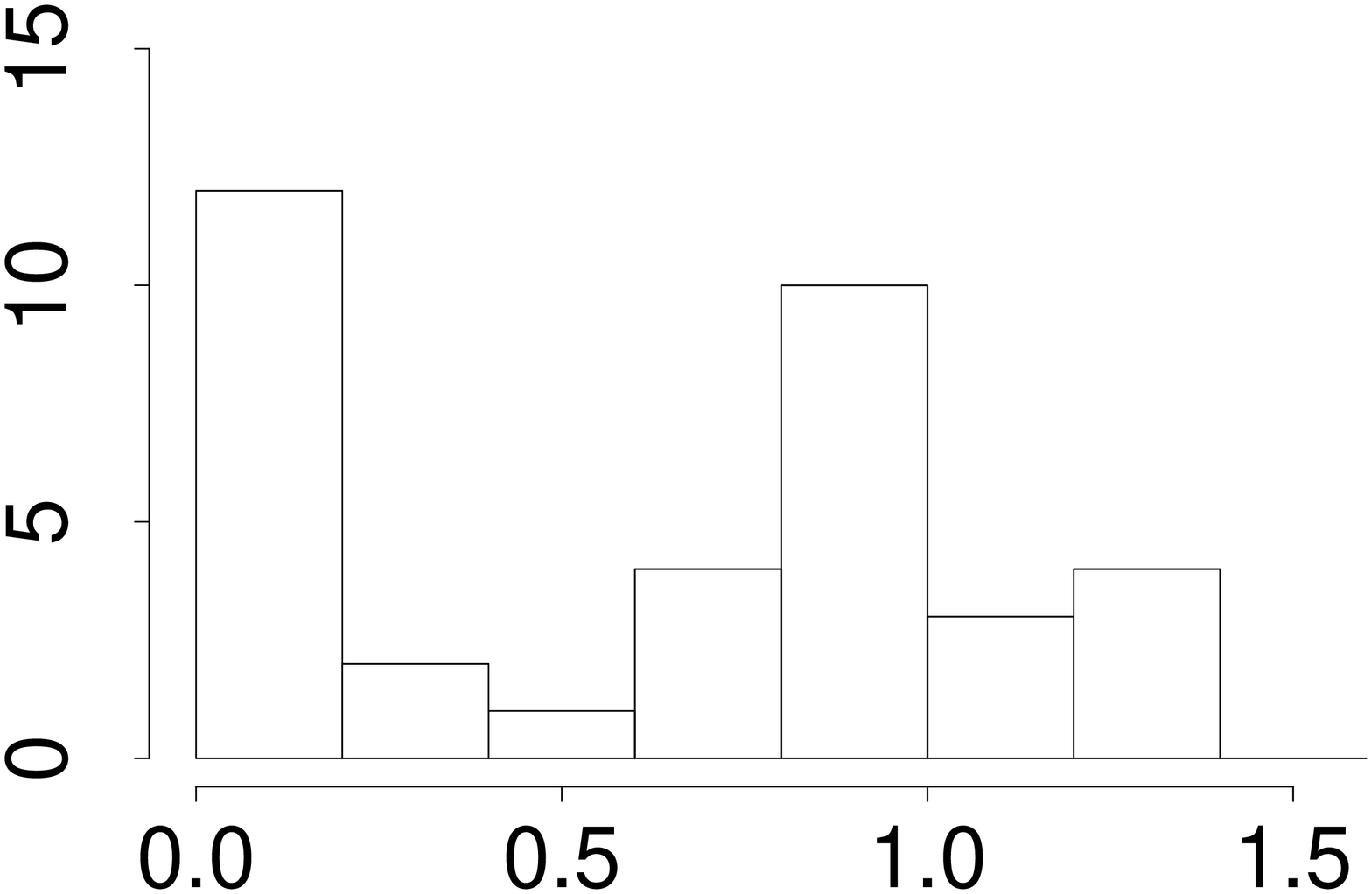}}
      \put(-2,36){\begin{sideways}\small frequency\end{sideways}}
      \put(55,3){\small fitness}
    \end{picture}
  }
  \subfigure[\label{fig:fitnessAnalysis:ahhs2}AHHS2]{
    \begin{picture}(120,80)
      \put(10,95){\includegraphics[angle=270,width=120\unitlength]
        {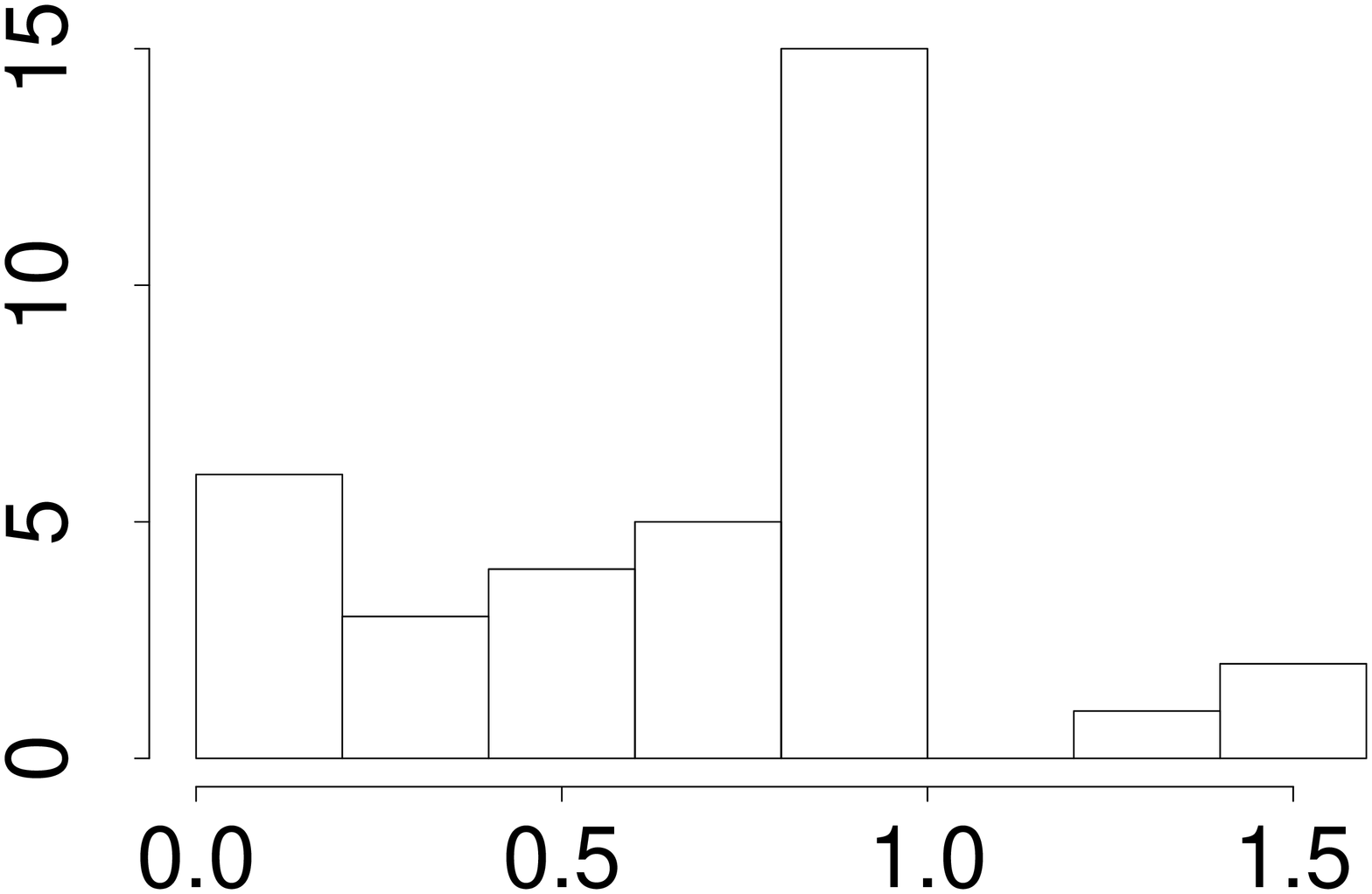}}
      \put(-2,36){\begin{sideways}\small frequency\end{sideways}}
      \put(55,3){\small fitness}
    \end{picture}
  }
  \caption{\label{fig:fitnessAnalysis}Fitness landscape neighborhood,
    fitness histogram of 35 samples of mutated controllers, fitness of
    the original controller is for AHHS1: 0.84, for AHHS2: 0.81.}
\end{figure}

\section{Conclusion and Outlook}

We have reported the application of our hormone control approach to
the domain of evolutionary modular robotics. The automatic synthesis
of controllers, that facilitate locomotion of organisms built from
five robot modules, has been effective in a majority of the
evolutionary runs. Almost all evolved controllers are able to generate
a form of locomotion that takes the organism at least to the
wall. A~majority of the evolved controllers were able to overcome the
wall. An unexpected vast diversity of locomotion paradigms was evolved
especially in the second version of the simulation. On the one hand,
this shows the complexity of the gait learning task in modular
robotics because there are many solutions of similar utility. On the
other hand, it shows the diversity of behaviors representable by AHHS
controllers.

Whether the redesigned controller AHHS2 is generally superior to the
original AHHS1 design is still an open question. However, in case of
the inverted pendulum it performs significantly better. In the gait
learning scenario AHHS2 shows a higher diversity and behaviors with
smoother movements resulting in more reliable locomotion.

There are many open issues and this research track is rather at its
beginning. Our future research will include the following. The
different possibilities of initializations need to be investigated
extensively. For example, the controllers could be initialized with
specialized sensor, hormone, and actuator rules (i.e., weights of
1). Scalability and more complex tasks from the domain of modular
robotics will be investigated (e.g., organisms with more modules). We
plan to use environmental incremental evolution (e.g., steadily
increasing heights of walls) as reported by \citet{nakamura00}. The
dynamic adaptation of rule numbers by evolution will be
investigated. Hence, we will evolve hormone reaction networks through
complexification similar to \citep{stanley04}. Finally, we plan to
check the controllers' exploration of the sensory-motor space,
especially, during the initial generations to get a better
understanding of what facilitates a high diversity of solutions.

\section{Acknowledgments}

This work is supported by: EU-IST-FET project `SYMBRION', no.~216342;
EU-ICT project `REPLICATOR', no. 216240.

\footnotesize
\bibliographystyle{apalike}
\bibliography{merged}

\end{document}